# THE PARTICLE TRACK RECONSTRUCTION BASED ON DEEP NEURAL NETWORKS

*Dmitriy* Baranov[1], *Sergey* Mitsyn[1], *Pavel* Goncharov[2,*] and *Gennady* Ososkov[1]

[1]Joint Institute for Nuclear Research, 6 Joliot-Curie street, Dubna, Moscow region, Russia
[2]Sukhoi State Technical University of Gomel, October Ave, 48, Gomel, Republic of Belarus

**Abstract.** One of the most important problems of data processing in high energy and nuclear physics is the event reconstruction. Its main part is the track reconstruction procedure which consists in looking for all tracks that elementary particles leave when they pass through a detector among a huge number of points, so-called hits, produced when flying particles fire detector coordinate planes. Unfortunately, the tracking is seriously impeded by the famous shortcoming of multiwired, strip in GEM detectors due to the appearance in them a lot of fake hits caused by extra spurious crossings of fired strips. Since the number of those fakes is several orders of magnitude greater than for true hits, one faces with the quite serious difficulty to unravel possible track-candidates via true hits ignoring fakes. On the basis of our previous two-stage approach based on hits preprocessing using directed K-d tree search followed by a deep neural classifier we introduce here two new tracking algorithms. Both algorithms combine those two stages in one while using different types of deep neural nets. We show that both proposed deep networks do not require any special preprocessing stage, are more accurate, faster and can be easier parallelized. Preliminary results of our new approaches for simulated events are presented.

## 1 Introduction

One of significant parts of the event reconstruction in particles track detectors in modern high energy and nuclear physics (HENP) is the tracking procedure that consists in finding tracks among a great number of so-called hits produced by flying particle interaction with sequential coordinate planes of tracking detectors. This procedure is especially difficult for modern HENP experiments with heavy ions where detectors register events with very high multiplicity.

Besides, while working on BM@N experiment [1] we have faced with the famous shortcoming of GEM strip detectors when a great amount of fake hits appears along with real hits because of extra spurious crossings of strips. The number of those fakes is greater for some order of magnitude than for true hits (see [2], for example).

A common method for dealing with track reconstruction is the combinatorial Kalman filter, which has been used with great success in HENP experiments for years [3]. However, the initialization of Kalman filter is a cumbersome process, because of a really vast search of hits needed to obtain so-called "seeds", i.e. initial approximations of track parameters of charged particles.

---

[*]e-mail: kaliostrogoblin3@gmail.com

There is a common opinion that machine learning algorithms could make a great contribution to the tracking problem due to their capability to model complex non-linear data dependencies. Keeping in mind the sequential nature of any tracking detector, we have proposed a two-step approach to the particle track reconstruction in BM@N GEM detector based on deep learning methods [2]. The first preprocessing step was devoted to a directed search of track-candidates. This procedure goes through all hit points at every detector station, starting from the first one and extending current track candidates by one hit to the next station also taking into account possible target coordinates. To speed up this preprocessing procedure, we took into account the direction of the magnetic field to arrange the search in two coordinate projections simultaneously. On YoZ projection, tracks are almost straight lines. Thus, merely by sorting hit indices array by y coordinate, making a confidence interval and executing a binary search, we can exclude all hits that are not in the confidence interval. For the XoZ projection, where the tracks look like circles, we enable a limited rotational component. The limitation is on the change in the rotation that should not change substantially. K-d tree data structure is used to bind the searching area of each continuation of track-candidates. After doing the first step of our tracking we obtain a bunch of track-candidates, which should be divided into two groups: real tracks and, so named, ghost tracks formed by fakes and, possibly, by parts of different tracks. To addressing the classification task, we utilized a deep recurrent neural network (RNN) as a classifier. The best results of validation efficiency were obtained by the combination of one convolutional layer and two GRU layers [4] one after another. As the result, we reach the testing efficiency 97.5%. Trained RNN can process 6500 track-candidates in one second on the single Nvidia Tesla M60. The full algorithm description one can find in [2].

In the given work we improve two-step tracking model in order to overcome its disadvantages by combining two-steps in one end-to-end trainable deep learning network. We propose two different approaches to realize our idea. The first approach, although it remains sequential, due to a significant change in the structure of the neural network and its loss function it combines both stages in one end-to-end pass.

The second approach overcomes the sequential nature of the tracking algorithm since it uses a completely different neural net as a type of a YOLO-like [5] convolutional neural network with CoordConv layer [6].

Both new approaches do not need any preprocessing, much better suitable to be parallelized and should outperform their progenitor in speed and accuracy.

## 2 Sequential tracking with one RNN

The previous two-step tracking does not have a suitable performance due to spending a lot of time while building a K-d tree structure for every event always from scratch. Thus, we decide to come up with a neural net model, that can predict both the probability whether or not they belong to a true track, and the searching area for the continuation of track-candidate on the next coordinate plane based on the input set of points.

**Model architecture.** To build a new model we changed considerably our deep RNN classifier from its structure described in [2] by replacing the bidirectional GRU layer with the regular one-directional (it reduces the number of neurons by half in every recurrent layer), by removing the dropout layers and, eventually, by adding to the output layer a special regression part needed to predict an elliptical area intended to search for the track-candidate continuation on the next coordinate plane.

This regression part consists of four neurons, two of which with linear activation function determine the ellipse center and another two – define the semiaxes of that ellipse.

Neurons, which predict the semiaxes of ellipses have softplus activation [7]. Our new model, presented on figure 1, takes, as input, sets with the different number of hits in them: from two (target and zero station's hit) up to N, where N is the number of detector readout stations. For N=2 the output will be presented by the regression part only, whilst for the sequences of the maximum length N, the output consists of the only one sigmoid neuron, since we do not need to seek for the next point on the very last station.

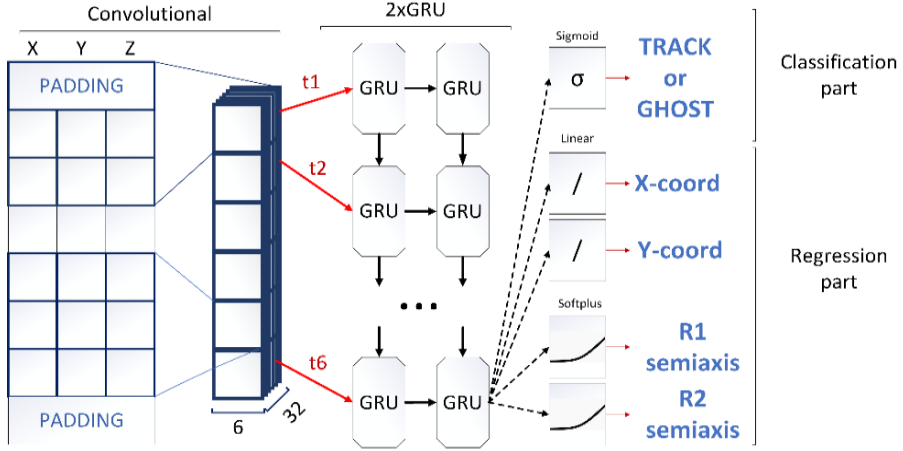

**Fig. 1.** The scheme of the deep RNN for tracking.

**Cost function.** To train the proposed new model we introduce an ad hoc cost function, defined as follows:

$$J = max(\lambda_1, 1-p)\, FL(p,p',\alpha,\gamma) + p\left(\lambda_2\sqrt{\left(\frac{x-x'}{R1}\right)^2 + \left(\frac{y-y'}{R2}\right)^2} + \lambda_3 R1 R2\right), \quad (1)$$

where $\lambda_{1-3}$ are the weights for each part of equation; $p$ is the label that indicates whether or not the set of input points belongs to true track; $p'$ is the probability of track/ghost to be predicted by network; $x'$, $y'$ are coordinates of the ellipse center, predicted by network; $x$, $y$ are coordinates of the next point of the true track segment; $R1$, $R2$ are semiaxes of the ellipse; $FL(p,p',\alpha,\gamma) = -p\alpha(1-p')^\gamma \log p' - (p')^\gamma(1-p)(1-\alpha)\log(1-p')$ is a balanced focal loss [8] with a weighting factor $\alpha \in [0,1]$. Introduction of such a weighting factor is the common method for addressing any class imbalance. We set $\alpha = 0.95$. The idea of the focal loss function is to down-weight well-classified examples and thus focus training on hard negatives. The focusing parameter $\gamma$ (we set it to 2) smoothly adjusts the rate at which easy examples are down-weighted.

While dealing with fake tracks we do not need to consider the error attended to ellipse size and position, for that reason we put the label $p$ before the second part of the equation. Thus, when the label is 0 (= ghost), we multiply the error of the regression part by zero. And, on the contrary, the $max(\lambda_1, 1-p)$ is utilized to focus on the classification error, when the regression part is turned off.

Summing up briefly all above mentioned, the equation (1) consists of three parts corresponding to $\lambda_{1-3}$: classification loss, point in ellipse loss, ellipse size penalization.

**Dataset and training setup.** To compare results we use the same dataset, prepared in [2], for two-stage tracking for training and evaluating the new model, namely, 15 thousand events with 20-30 tracks per event were simulated and labeled then by K-d tree search for obtaining the bunch of labeled track-candidates (seeds). Eventually, we obtained 82 677 real tracks and 695 887 ghosts. During every training iteration, the training set of seeds is split into three groups of track-segments containing a different number of points (from 2 to 6). For each of these seeds, the network should predict the probability whether that set of points belongs to a true track (except 2 points) and also an elliptic area, where to search for a track continuation (except 6 points). Deep RNN has been trained with $\lambda_1 = 0.5$, $\lambda_2 = 0.35$, $\lambda_3 = 0.15$, $\alpha = 0.95$, $\gamma = 2$ for 50 epochs with batch size equals to 128 and Adam optimization method. The loss value on the test subset of dataset after training stabilized on 0.019.

**Results** of the model evaluation on the test subset of data (250K of seeds with the factor of 1:10 – one true track opposite to ten ghosts) are presented in table 1. We have leveraged several metrics to test the model correctness to predict belonging of inputs to true tracks. They are known in statistics as accuracy, precision, and recall [9].

**Table 1.** Results of applying several metrics to the trained network for different input length.

|           | 3 points | 4 points | 5 points |
|-----------|----------|----------|----------|
| **Recall**    | 98.2%    | 99.0%    | 98.3%    |
| **Precision** | 49.0%    | 57.0%    | 70.0%    |
| **Accuracy**  | 88.0%    | 92.0%    | 95.2%    |

Accuracy (also known as efficiency) is the fraction of predictions our model got right, but it becomes useless while dealing with an imbalanced dataset. Therefore, precision and recall metrics are used in these cases as more informative. Precision tells us how many of the objects classified as true tracks were correct. Recall means how many of the objects that should be marked as true tracks were actually selected. Thus, recall expresses the ability to find all true tracks in a dataset, while precision expresses the proportion of data, our model says was true, actually were true tracks.

In our study, we speed up significantly our deep network calculations by using multicore computational systems and multiprocessor graphics cards or GPUs via the facilities provided by the JINR supercomputer GOVORUN [10]. The speed of the test run has reached 3 483 608 track-candidate/sec on 2x Tesla V100.

## 3 Look Once On Tracking

As it was mentioned above, all existing algorithms for tracking are sequential. Also, most of them do not see the whole picture, they are limited only to individual trajectories pass while trying to ignore the neighbors. Instead, we unify the separate components of tracking into a single YOLO-like convolutional neural network [5]. Our network consumes the whole event as, if it is a picture, but instead of RGB channels we replace them by discretized contents of detector stations considering each of them as a channel in depth. In this way, we obtain the input tensor which depth equal to the number of stations, its width and height depend on the resolution of the detector, i.e. the size of the biggest station. This means our model represents the whole event with all tracks and noisy hits in it on the global level.

We named our model LOOT, which means «**L**ook **O**nce **O**n **T**racks». The design of the LOOT allows end-to-end training and obtaining high processing speed, because of avoiding of the sequential problems while maintaining high average precision. Every coordinate plane for every station is divided into a grid with the size of the biggest station depending on the selected resolution. Each grid cell may have in it either a hit the value of such cell equals to 1, otherwise 0. Withal, hit may be a fake.

Each grid cell predicts X and Y shifts from the current position on the grid to every next point of the track together with the probability that there is a true track in a cell called confidence score. This means our network predicts 11 values for the tracks registered by 6 stations and the confidence score value. In more details, 5 X-axis shifts and 5 Y-axis shifts with respect to the corresponding cell indexes ij, where shifts are simple indents, for instance, if the current hit's position is in the point (2, 3) and the next is in the (4, 1), then X-shift will be 2 and the Y-shift will be -2. The confidence score is a probability of the grid cell to be the point of some track. If no track exists in that cell, the confidence score should be near zero. Thus, each grid cell can include only one track.

During the testing, we set to zero all cells which confidence score is below some threshold (say, 0.5), and then use it as a mask, multiplying by the predicted shifts.

**Model architecture.** Our experiments showed that such kind of model like LOOT cannot directly predict the pixel shifts while maintaining high average precision. At first, we thought that the problem belongs to the mean-squared error loss, which strongly averaging all outputs. But then we found that it is a very common problem named *coordinate transform problem* [6]. To address this problem, we add a special CoordConv layer proposed by the authors in [6]. Coordinate convolution is an operation that includes concatenation of the additional information of the row and column indices of grid cells to the input tensor and then applying 1x1 convolution to the expanded input.

After the CoordConv layer, we added 4 convolutional layers with ReLU activation and Batch Normalization, two of which extract 32 features and another two has 64 filters (see fig. 2). The output is a concatenation of the features extracted by two convolutional layers: the first one with the sigmoid activation function predicting the confidence score and the second one extracting $d \times 2 + 1$ feature maps, where $d$ is the number of stations.

Worse to note, that to avoid predicting tracks in empty cells, we tried to multiply the first station matrix to the shifts and confidence scores as an additional mask, but it didn't improve results at all. We also tried to create a model in a deep autoencoder manner, gradually squeezing the input tensor, since it was very useful in image recognition tasks [11], but it did not accelerate the speed of training and did not give higher efficiency.

The final output of our network is the w × h × [d x 2 – 1] tensor of predictions, see the fig. 2.

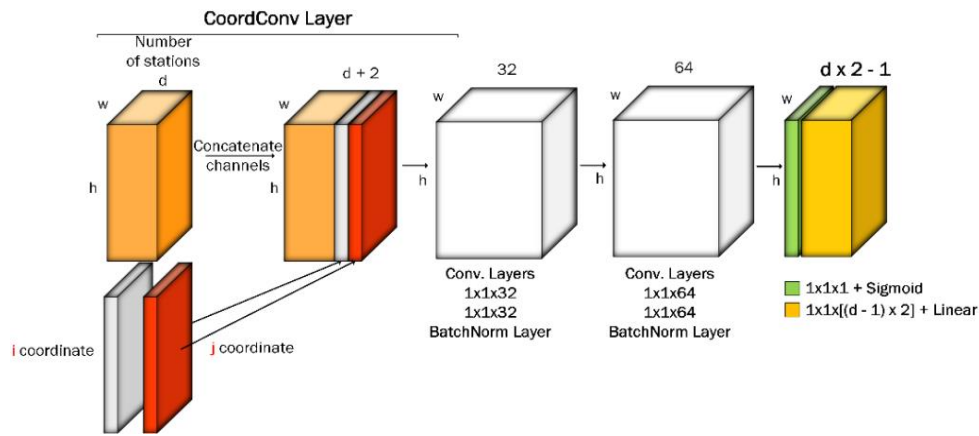

**Fig. 2.** The architecture of the LOOT model. As first layer we utilize a coordinate convolution layer followed then by 4 1x1 convolutional layers with ReLU activations doubling the number of input channels and the Batch Normalization layers too. Here **d** is a number of stations in the detector.

**Two-part custom objective.** To train the proposed new model we introduce an ad hoc loss function, defined as follows:

$$L = \frac{\sum_{i=0}^{w \times h} CE(C_i, \widehat{C}_i)}{w \times h} + \frac{\sum_{i=0}^{w \times h}(\text{Xshifts}_i * \widehat{C}_i - \widehat{\text{Xshifts}}_i)^2 + (\text{Yshifts}_i * \widehat{C}_i - \widehat{\text{Yshifts}}_i)^2}{nz}, (2)$$

where $CE(C_i, \widehat{C}_i)$ is a binary cross-entropy loss between predicted confidence scores – $C_i$, and real ones – $\widehat{C}_i$, $w \times h$ is the total number of grid cells, $nz$ – the number of nonzero elements. We avoid dividing by $w \times h$ because of a strong sparsity of the output matrix. Also, we multiply the shifts by the real confidence score, which is a matrix representing the hits from the first station, to omit the cells with fakes, they do not have to take part in shifts calculation.

Summing up briefly all above mentioned, the equation (2) consists of two parts: cross-entropy confidence loss and normalized squared error shift loss.

**Dataset and training setup.** We did not use that ready dataset which was used for two previous networks because it was elaborated for the concrete physical experiment with carbon-carbon collisions producing events with 15-20 tracks, but we were interested to try events with greater multiplicity. Therefore, we create a special event generator for the LOOT training and the proposed model evaluation, which produces a specified number of straight tracks, that is equivalent to events without magnetic field or can be attributed to particles with high momentum. Obtained hits coordinates are transferred then to the indices at a pixel grid cell with the given resolution.

For every training epoch, we generate 1000 batches with the size of 8, where each training sample is a tensor with 512x512x5 dimensions (5 stations with the size 512x512).

The number of tracks varies from sample to sample from the minimum number of 10 tracks and the maximum number of 100. To model fake hits, we sample positions of pixels from the uniform distribution, so our fake hits do not correlate each other unlike fakes in the GEM detectors. We generate fakes with the number twice greater than the number of true hits, i.e. for 100 tracks and 5 stations (500 true hits) one obtains 1000 uniformly distributed fake hits.

For the model validation, we generate 100 batches of the same size as training batches. As metrics for the evaluation, we use two functions: the first one computes the fraction of the true hits among all predicted hits, this corresponds to precision; the second metric, named efficiency is the percentage of tracks correctly found in relation to the number of simulated ones. To determine a well-detected track we, reconstruct events from the predicted output including shifts and probabilities, find nearest existing hits depending on the reconstructed from the shifts and count how many hits in the track coincide with hits of the corresponding monte-carlo track. We take a track as correctly found if it contains 70% or more of the points from simulated one.

We train the model for 20 epochs (160000 events) using Adam optimizer with default parameters.

**LOOT Limitations.** Presently, LOOT works only with tracks, which pass through all stations – it is a very rough assumption, consequence of this that we cannot detect tracks that have large scattering angles relative to the Z axis.

Also, we start predicting track hits based on the first point of the track and we suppose that every track starts from the first station, although in real life it is not always true. So, if the track would not be registered in the first station, our model will not recognize it.

We are still working to overcome the above-mentioned limitations.

**Results** of the model evaluation on the generated test data (1000 batches with 512x512 detector's resolution, number of tracks from 10 to 100 and fakes factor with the value of 2) are presented in table 2.

**Table 2.** Results of the evaluation of the trained network for a different number of tracks.

| # of tracks in event | 10 tracks | 50 tracks | 100 tracks |
|---|---|---|---|
| Rate o fake hits | 0.0009 | 0.0013 | 0.0043 |
| Efficiency | 99.4% | 98.8% | 97.97 |

Confidence threshold used to decide to drop out a track-candidate with the small confidence score is accomplished. That means, if the confidence score has a value smaller than the threshold, this grid cell with a track-candidate becomes empty. We set the confidence threshold value to 0.5.

As it was mentioned above, the tracking efficiency is a fraction of tracks with more than 70% hits detected rightly, among all tracks. One can see from table 2, that the obtained efficiency and fake hits rate look satisfactory.

The network model was made with the help of Keras library [12] using the TensorFlow [13] as backend.

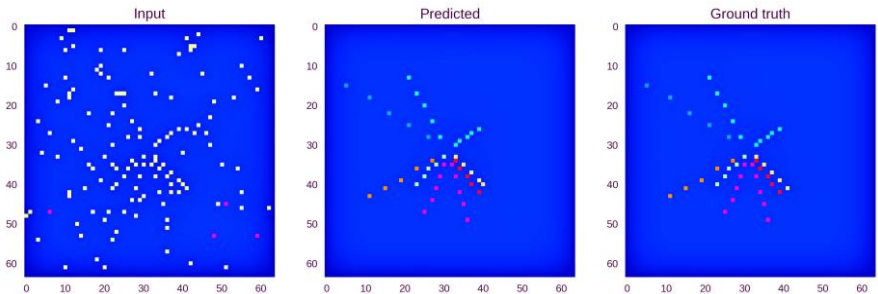

**Fig. 3.** Example of the model prediction. XoY plane of 3D event is shown. The left image is the LOOT input; the central image is the prediction and the right one – the target.

The example of the prediction of the trained model is illustrated on fig. 3, where XoY plane of 3D event is shown. We colored each track with different colors to make them better distinguished. Also, for the better visualization, we present here an event simulated with small detector resolution (64x64) and a small number of tracks.

In our study, we speed up significantly our deep network calculations by using multicore computational systems and multiprocessor graphics cards or GPUs via the facilities provided by Google Colaboratory [14]. The speed of the test run has reached 23 event/sec on a single Nvidia K80 with batch size equals to 100 and the detector's resolution – 512x512.

## 5 Conclusion

The two step-tracking algorithm, proposed in [2], was too slow because of the preprocessing procedure on the first stage based on the KD-tree 2D search. Therefore, we propose here two different approaches to overcome the disadvantages of the two step-tracking algorithm, by combining both steps in one end-to-end trainable deep learning network. The first approach, although it remains sequential, applies deep recurrent neural network with the special loss function as an alternative to Kalman filter with the ability to train the needed parameters from data. Nevertheless, it has the number of limitations, in particular, it cannot work fully end-to-end because each track is processed independently and there is no information about nearest tracks. In the second approach, a completely new neural network model is proposed. The LOOT network overcomes all mentioned problems of the previous models. It is fast, accurate and fully end-to-end trainable. Although there are some substantial questions of its adaptation to real events which are still waiting for their solution, we believe it opens a new horizon in the field of event reconstruction.

**Acknowledgments**: This work is partially supported by the State Committee on Science and Technology of the Republic of Belarus (contract No. 202/17(039/64) of 25/07/2017).